\documentclass[10pt,conference]{IEEEtran}
\makeatletter

\IEEEoverridecommandlockouts                              
                                                          
\usepackage{subfig}
\usepackage{tikz}
\usepackage{textcomp}
\usepackage{hyperref}
\usepackage{lipsum}
\usepackage{amsmath}
\usepackage{xspace}
\usepackage{balance}
\usepackage{graphicx}
\usepackage{xcolor,colortbl}
\usepackage{tabularx}
\usepackage{tabulary}
\usepackage{enumerate}

\DeclareMathOperator*{\argmax}{argmax}
\newcommand{\system}{DeComplex\xspace}

\begin{document}

\title{\system: Task planning from \\complex natural instructions by a collocating robot}

\author{\IEEEauthorblockN{Pradip Pramanick, Hrishav Bakul Barua, and Chayan Sarkar}
	\IEEEauthorblockA{TCS Research \& Innovation, India}
}

\maketitle
\thispagestyle{empty}
\pagestyle{empty}

\begin{abstract}
	As the number of robots in our daily surroundings like home, office, restaurants, factory floors, etc. are increasing rapidly, the development of natural human-robot interaction mechanism becomes more vital as it dictates the usability and acceptability of the robots. One of the valued features of such a cohabitant robot is that it performs tasks that are instructed in natural language. However, it is not trivial to execute the human intended tasks as natural language expressions can have large linguistic variations. Existing works assume either single task instruction is given to the robot at a time or there are multiple independent tasks in an instruction. However, complex task instructions composed of multiple inter-dependent tasks are not handled efficiently in the literature. There can be ordering dependency among the tasks, i.e., the tasks have to be executed in a certain order or there can be execution dependency, i.e., input parameter or execution of a task depends on the outcome of another task. Understanding such dependencies in a complex instruction is not trivial if an unconstrained natural language is allowed. In this work, we propose a method to find the intended order of execution of multiple inter-dependent tasks given in natural language instruction. Based on our experiment, we show that our system is very accurate in generating a viable execution plan from a complex instruction.

\end{abstract}

\section{INTRODUCTION}
Recent developments in robotics have enabled robots to move from the secluded industrial setup to our daily surroundings. These cohabitant robots are used in homes or hospitals as caregiver, helper, companion, in the factory floors or offices as coworkers~\cite{pramanick2018defatigue}, assistant~\cite{martinez2019pharos}, etc. These applications not only broaden the scope but also necessitate more frequent interaction between a human and a robot. Instructing a robot in natural language adds to its usability in such dynamic environments. While the primary focus of these robotic systems should be improving the accuracy of the model(s) that predict(s) the meaning of the instruction, the predominant trend of doing so is to impose various constraints on the space of linguistic variations, ambiguity, and complexity of the language. 

\textbf{Motivation}. Natural human-robot interaction requires that a non-expert user should be allowed to instruct the robot in a flexible way that suits his/her need. One such flexibility is conveying many tasks at once while instructing the robot~\cite{shah2018follownet,nicolescu2019learning}. This is particularly convenient for the human if the area where the robot is operating is large and therefore not fully observable to the user and/or the robot works in shared autonomy where it has to work autonomously after instructed, possibly in a different room/area. This is also the case when the robot is being teleoperated by a remote user. In such situations, the instruction can be given to check the facts that the user is unsure of and also provide alternative tasks in case of failure and/or undesired outcomes. Such complex instructions can have many inter-dependencies between the tasks, which the robot needs to understand to perform them as intended.  
\begin{figure}
	\centering
	\includegraphics[width=\linewidth]{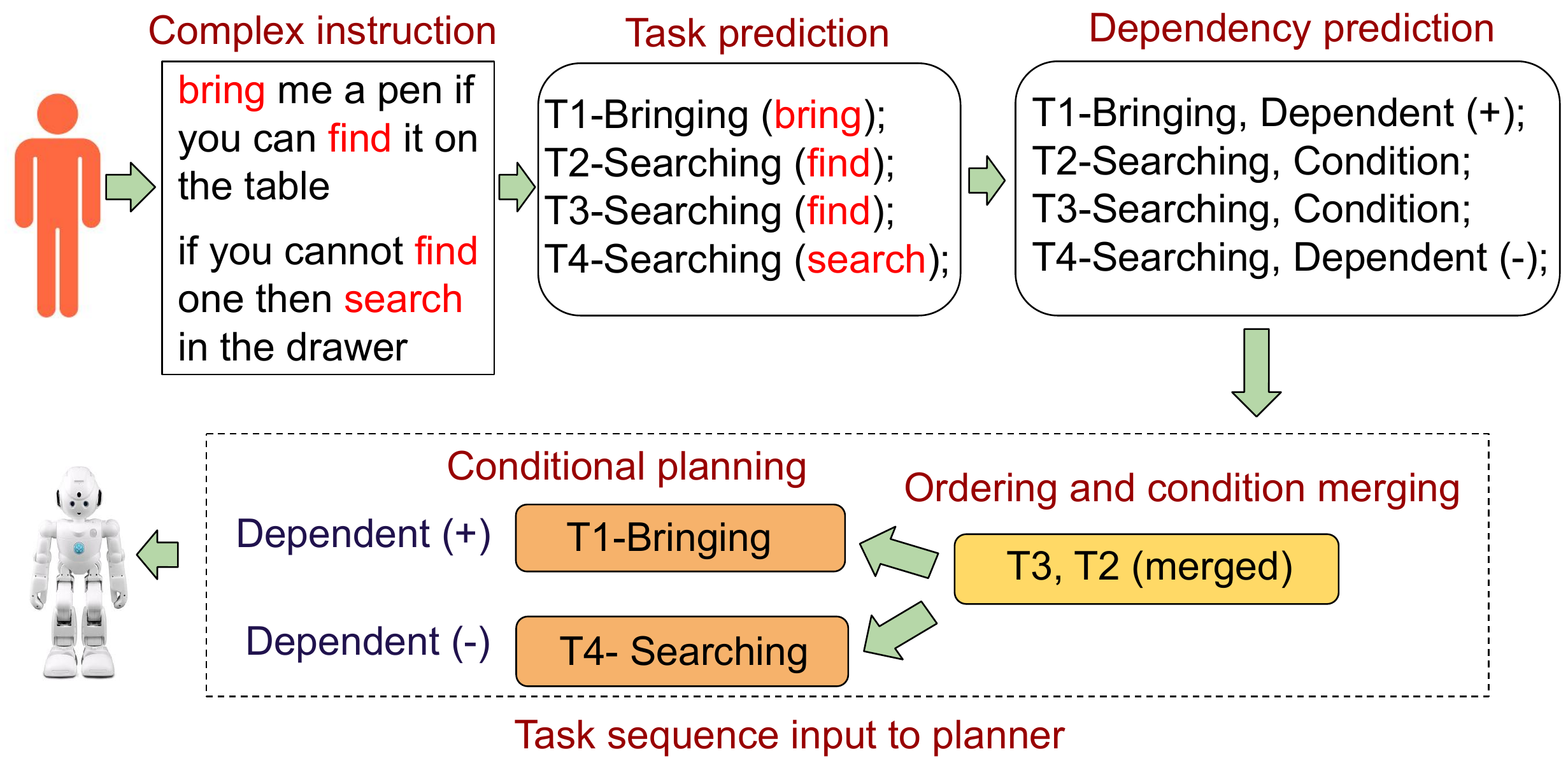}
	\caption{A high-level pipeline to generate a viable execution plan from complex natural language instruction.}
	\label{fig:pipeline}
\end{figure}

\textbf{Problem description.} Existing works that translate natural language instruction to a sequence of actions, either constrain the instruction to a single task~\cite{lu2017integrating,paul2018temporal} or assume multiple tasks are performed sequentially~\cite{matuszek2013learning,tellex2011understanding,pramanick2019enabling}. In the later case, the sequence is assumed to be the order in which the tasks appear in the instruction. However, this assumption may not hold if natural language input is to be assumed. For example, in the instruction \textit{``bring me a pen if you can find it on the table, if you cannot find one then search in the drawer''}, the robot has to find the pen first, before attempting to bring it, although the bringing task appears in the instruction earlier. Moreover, the execution of a task may depend upon a condition or the outcome of another task. In the previous example, both the tasks of bringing the pen and the task of searching the drawer is dependent upon whether the pen is on the table or not. 

\textbf{Approach.}
In this work, we present a system, called \system to understand complex natural language instruction (Fig.~\ref{fig:pipeline}) and generate a viable execution plan. We consider complex instructions that are composed of multiple tasks given as both commands and statements and the task execution is constrained by ordering and/or execution dependency. We define ordering dependency as the case when the tasks present in an instruction must be ordered in a certain way to reach the human desired goal. By execution dependency, we refer to the scenario when the execution of some tasks is dependent upon the outcome of other tasks. Please note we do not assume a complex instruction all the time. So, at first, our classifier inspects if a given instruction is a complex instruction composed of multiple tasks and whether there is any dependency at all. After identifying the tasks conveyed by the instruction and the dependencies involved between the tasks, our system identifies the corresponding parameters for each of these tasks given in the instruction and finally generates a viable plan for the instruction that a robot can execute. In case the tasks have execution dependency, i.e., the output of one task feds into another, we generate a conditional (viable) plan, which can follow a particular execution path based on the situation. 

Although there is an inherent ordering dependency of selecting the appropriate sequence of actions to execute a certain task and we solve that using a planner. In this work, we are more interested in understanding the dependencies among the tasks themselves that are explicitly or implicitly provided by the human through natural language instruction. 
Specifically, our major contributions are two-fold.
\begin{itemize}[]
	\item From complex instruction in natural language, we find the set of tasks, inter-dependency among them if any, and extract input parameters for each task.
	\item We merge the duplicate tasks and create a control flow graph of the unique tasks considering causality and inter-dependency among them so that a viable (high-level) execution plan can be generated.
\end{itemize}

\section{RELATED WORK}
\label{sec:related}
A robot's capability to understand natural language instructions is limited by many factors. One of them being the inability to understand the flow of actions and their effects in a complex set of instructions (such as having conditional statements/sentences in instructions). Our work primarily aims at proposing a strategy to handle such complex instructions with efficacy and allowing the robot to act accordingly. This section puts forward some of the related works in this respect.
\subsection{Task planning from instruction}
Task planning for robots from natural language instructions has been receiving a lot of attention in recent years. The predominant approach includes understanding the task and its arguments from a parsed semantic representation, followed by mapping actions to world state~\cite{tellex2011understanding,thomas2012roboframenet}, planning using post-conditions~\cite{misra2015environment,antunes2016human,lu2017integrating,pramanick2019enabling} or using rich knowledge-bases that includes task to action decomposition information~\cite{chen2013handling,nyga2018grounding}. To tackle the ambiguity and incompleteness of natural language, dialogue agents have been proposed ~\cite{thomason2015learning,thomason2019improving,pramanick2019strategy}. Alternatively, techniques of directly learning of task plans from natural language instruction have been explored in~\cite{misra2015environment,tellex2011understanding}. In this work we follow the approach of task understanding described in our previous work~\cite{pramanick2019enabling}. However, one of the major limitations we solve in this work is that the existing techniques for task understanding cannot handle complex instruction, specifically if the tasks are inter-dependent. 
\subsection{Understanding complex instruction}
Many existing approaches consider a single task per instruction or assume the tasks are independent and can be planned together by satisfying a conjunction of post-conditions or goals~\cite{thomas2012roboframenet,lu2017integrating,chen2013handling}. Others assume that multiple tasks are serialized and they generate the plan by independently solving the planning problems for each task while considering the changes in world state for the preceding tasks~\cite{pramanick2019enabling,tellex2011understanding,kollar2014grounding}. Whereas, some early approaches to understand the ordering of multiple tasks focuses on finding out tasks that are to be performed multiple times until some condition is satisfied~\cite{kollar2010toward,matuszek2013learning}. However, they do not consider execution dependency, out of order appearance of tasks and they overlook many challenges by restricting to only navigational instructions. Other approaches to understand execution dependency imposes constraints on the language, i.e., they only allow structured English~\cite{kress2008translating,mericcli2014interactive,dzifcak2009and}. 

Although understanding complex instructions given to a robot has not been studied in depth, there are some existing works that includes this feature. For interactive task learning, usage of complex instructions has been explored, but assuming structured or constrained language specifications. Finucane \textit{et al.}~\cite{finucane2010ltlmop} and Chai \textit{et al.}~\cite{chai2017teaching} proposed rule-based methods for parsing complex action specifications to control logic. Cantell \textit{et al.} explored parsing complex instructions into pre-conditions, post-conditions and actions for specifying planning operators~\cite{cantrell2012tell}. In~\cite{boteanu2016model}, natural language commands contains a single task, whereas complex instruction is only allowed for providing action specifications that also in a structured language. 

The existing approaches that understand and generate task plan from complex instructions while allowing natural language~\cite{nicolescu2019learning, misra2015environment,pomarlan2017natural} use semantic parsers augmented with rules, which can't handle unseen linguistic variations. Other approaches embed planning for multiple tasks in end-to-end training~\cite{artzi2013weakly,misra2016tell}. However, such techniques of direct training with environment-specific plans do not generalize to novel situations, and a significant annotation effort is required to introduce a new task. In contrast, we propose a probabilistic graphical model for understanding complex instructions that need a resolution of interdependencies, which can generalize better. We use a data-driven approach to predict task dependency and deterministically re-order the tasks using the prediction, which is followed by task planning using post-conditions. This aids our system to understand the dependency of a task even if its appearance in the instruction is unusual or it is given in a separate sentence.

\begin{figure*}
	\centering
	\includegraphics[width=0.95\linewidth]{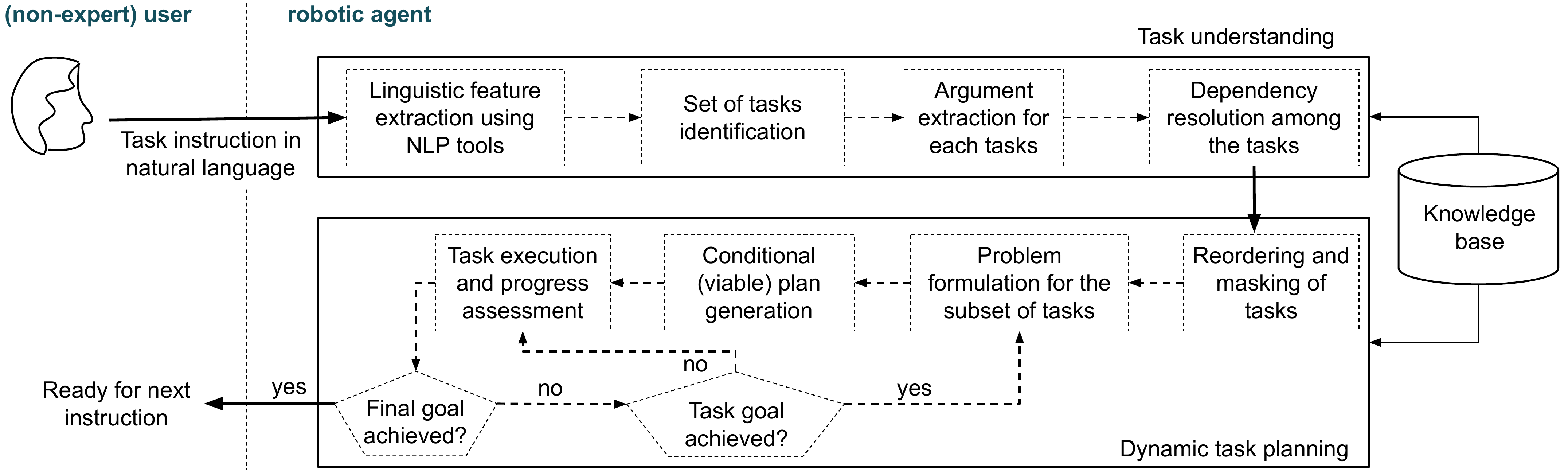}
	\caption{System overview of \system for task plan generation from complex natural instructions.}
	\label{fig:overview}
\end{figure*}

\section{\system IN DETAILS}
\label{sec:overview}
Fig.~\ref{fig:overview} shows an overview of \system that consists of -- (i) a \textit{task understanding} component and (ii)~a \textit{dynamic task planning} component. The first component parses a natural language instruction and extracts various linguistic features, identifies the set of tasks present in the complex instruction, identifies the dependencies among the tasks, and extracts the arguments for each type of task. On the other hand, dynamic task planning first determines the action sequence of the tasks by matching their post-conditions and inter-dependencies and generate a viable plan for the robot to execute. \system consults a knowledge-base that contains question templates for dialogue, pre- and post-condition templates for plan generation, and a world model represented by logical atoms. The system model is adapted from our previous work called ``task conversational agent for robots (TCAR)"~\cite{pramanick2019enabling}. However, as mentioned previously that TCAR cannot handle complex instructions and task dependency, it would fail to generate a plan for such instruction. In the following, we discuss the major building blocks of \system in details.

\subsection{Task set identification}
To execute the instruction, first, the robot has to identify the set of tasks contained in the instruction. We tokenize the instruction first and then mark a token as a task if it has an unambiguous goal that the robot knows to achieve by task planning. Given an instruction $I$ as a sequence of tokens, $I=\{w_1,w_2,\dots, w_n\}$, we model the identification of tasks as a sequence labeling problem and predict the task labels $t_{i:n}$. We solve this using probabilistic graphical model, called a Conditional Random Field (CRF). This includes marking tokens in commands such as ``go to the kitchen'' as well as in statements ``if the coffee is hot'', as tasks. We identify the tasks conveyed in instruction by jointly marking the tokens that denote a task and by determining the type of the task. We label the tasks from the set of task types known to the robot. All the non-task tokens are labeled as $O$. As an example, for given instruction \textit{``if the coffee is hot, bring it to me''} the CRF model labels the tokens as:

\textit{ \{ {\small if - O, the - O, coffee - O, is - check\_state, hot - O, bring - bringing, it - O, to - O, me - O} \} }.

We also use the same CRF to add new task types and new feature functions to appropriately predict the tasks from statements, which is not considered in the existing work. 

\subsection{Argument identification}
Each task is associated with one or more arguments and the argument values are grounded entities and states in the environment using a knowledge base. We also model the argument prediction from an instruction as sequence labeling using a CRF, by predicting the argument labels appended by the \textit{BIO} notation, i.e., we mark each token $a_i$ as inside (I), outside (O) or as beginning (B) of the argument label. For this, we extend the model described in~\cite{pramanick2019strategy}. Let $T$ be the set of predicted task types after removing non-task tokens, then given the task prediction labels $t_{i:n}$, the CRF for argument prediction estimates the following conditional probability distribution, 
\vspace{-0.3cm}
\begin{align*}
P(a_{1:n}|w_{1:n})= \alpha \exp \bigg \{\sum_{i=0}^n \big (\sum_{j=0}^{l-1} \lambda_j f_j(I,i,a_{i-1},a_i)  \\+\lambda_{l}\; g(I,i,t_{1:n}) \big ) \bigg \},
\end{align*}
where there are $l-1$ arbitrary feature functions and $\lambda_j$ is the learned weight of the j\textsuperscript{th} feature function, and $g$ with the weight $\lambda_l$ is used to associate a task type label with each word. The function $g$ is defined as the following,
\[
g(I,i,t_{1:n}) = 
\begin{cases}
\phi,& \text{if } t_i \in T\\
t_j,& \text{else if } t_j \notin T \text{ and } j>i.
\end{cases}
\]
This means the task association feature function ($g$) associates every token followed by a task to be its target until it finds a new task. Although, the arguments can be predicted without this feature, using this feature helps classification because the model can learn to predict the argument type only considering the arguments that are possible for the task type. This simple assumption works well for instructions containing a single task, or instructions consisting of only imperative sentences or commands. However, in a complex instruction, the target tokens for predicting argument can both precede and succeed the task. For example in the following instruction: \textit{``Bring me some coffee if it is hot''}, the \textit{bringing} task should be associated with the succeeding token `coffee' and the \textit{check\_state} task should be associated with both the preceding token `it', and the succeeding token, `hot'. 
Thus, during inference, using this feature function naively can give inaccurate associations.

We solve this feature association problem by introducing a new feature function that maximizes the likelihood of predicted arguments. To do this, the function generates a set of different task association features for a single inference for the tokens in-between two tasks. Then it makes multiple predictions and calculates the joint probabilities of each predicted label sequence. Finally, it chooses the prediction with the maximum joint probability, i.e.,
\vspace{-0.1cm}
\begin{align*}
\argmax_{} \bigg ( \prod_{j=1}^{k} P(a_{1:j}|w_{1:j},t_1), P(a_{j+1:k}|w_{j+1:k},t_2)\bigg ),
\end{align*}
where there are $k$ tokens in between the tasks $t_1$ and $t_2$.

For the previous example of complex instruction, our CRF model for argument prediction labels the token sequence as:

\textit{ \{ {\small if - O, the - B-Object, coffee - I-Object, is - O, hot - B-State, bring - O, it - B-Object, to - B-Goal, me - I-Goal} \} }.

After extracting the arguments, we use a co-reference resolver to replace anaphoric references using pronouns such as \textit{it, them} etc. with the corresponding arguments (nouns) of the preceding tasks.

\subsection{Task dependency resolution}
If there are multiple tasks present in the instruction as reported by the task identification stage, we resolve their dependencies by predicting if the execution of a task is dependent on the execution of another task and if so, we also predict the nature of the dependency. We jointly model these two predictions as a sequence labeling problem by predicting task dependency labels for a sequence of predicted tasks. Given the instruction, $I$ and the corresponding sequence of task type labels, $T_p=\{t_1,t_2, \dots, t_n \}$, we predict the sequence of task dependency labels, $D=\{d_1,d_2, \dots, d_m \}$, only for the known task type labels, i.e for $t_i \neq O$. Table~\ref{tab:task-dependency-lables} shows the task dependency labels and their definitions. 

We perform this sequence labeling using a CRF. As we are only interested in predicting the dependency label of a token that has been marked as a task by the task identification model, we consider only such tokens for the prediction. As an example of the labeling, the task sequence in the previous instruction is labeled as:

\textit{\{ {\small check\_state - conditional, bringing - dependent\_positive}} \}

\begin{table}[t]
	\centering
	\caption{Task dependency labels predicted by CRF.}
	\begin{tabular}{|p{1.7cm}|p{5.7cm}|}
		\hline
		\textbf{Label} & \textbf{Definition} \\ \hline
		conditional & The task has one or more dependent tasks. \\ \hline
		dependent-positive & The task should be executed if the preceding \textit{conditional} task yields the desired outcome. \\ \hline
		dependent-negative & The task should be executed if the preceding \textit{conditional} task fails or yields a undesired outcome. \\ \hline
		sequential & The task is not explicitly dependent upon another task and the order of execution is assumed to be corresponding to its position in the instruction. \\ \hline
	\end{tabular}
	\label{tab:task-dependency-lables}
\end{table}
In the following, we discuss the challenges in understanding these dependencies between tasks from a complex instruction and our strategies to solve them.
\subsubsection{Challenges} 
The difficulty in understanding the dependencies between tasks arises from several intricacies of natural language. We noted three major challenges in this regard: 
\begin{itemize}
	\item Unmarked dependency: We say a task has a marked dependency when there is token preceding the verb that can determine the task's dependency type. For example, consider the instruction: \textit{``If you can't find it on the table, look in the cupboard''}, although the first searching task in the sub-ordinate clause has a dependency marked by the token \textit{`If'}, the second (searching) task in the independent clause has no lexical element that can determine its dependency. In other words, if the independent clause \textit{``look in the cupboard''} is inspected separately, the task seems to have no dependency at all, which makes its prediction non-trivial.
	\item Out of order appearance: A prerequisite\footnote{If task T1 is dependent on task T2, then T2 is prerequisite of T1.} is usually followed by one or more dependent tasks and this co-relation is useful for building rules for understanding task dependencies~\cite{misra2015environment}. However, it is also natural to convey a dependent task, followed by conditional prerequisite. As an example, consider the instruction \textit{``Bring me a pen, if you find one on the table.''} where the bringing task is dependent on finding it first, but the prerequisite is stated later. Furthermore, such out of order, dependent tasks usually have unmarked dependencies that are already difficult to predict. 
	\item Implicit dependency: We say a task has an explicit dependency if all of its dependent tasks appear in the same sentence. If the prerequisite of a task appears in a different sentence, we call it an implicit dependency. For example, in the instruction: \textit{``Turn on the tv. If you cannot, bring me the remote.''}, the task of bringing in the 2nd sentence is dependent upon a task in the first. Implicit dependencies are difficult to predict because the prerequisites usually have unmarked dependencies and as the tasks appear in different sentences, syntactic relations between the two tasks can't be found, which is otherwise useful for the prediction.
\end{itemize}
\subsubsection{Model}
We use a linear-chain CRF model for predicting the task dependency labels in an instruction marked with task types. The CRF is a factor graph, that predicts a label sequence, given an observation sequence. The CRF model for predicting the task dependencies estimates the following conditional probability of a label sequence $d_{1:m}$, given the sequence of tasks $t_{1:m}$ and the token sequence, $w_{1:n}$,
\begin{flalign*}
P(d_{1:m}|t_{1:m}, w_{1:n})= \\ \alpha \exp \bigg \{\sum_{i=0}^m \sum_{j=0}^k \lambda_j f_j(w_{1:n},t_i,d_{i-1},d_i)  \bigg \},
\end{flalign*}

where  $\alpha$ is a normalization factor, $f_j$ is the $j^{th}$ arbitrary feature function, $\lambda_j$ is the weight of the $j^{th}$ feature function, and k is the number of such feature functions. Each feature function $f_j$ is defined over the token sequence, the task type, and two consecutive labels. We use several grammatical features that include parts of speech (POS) tag and dependency parse tree\footnote{Dependency parse tree is a grammatical structure of a sentence, different from our terminology of task dependency.}. We extract the features using a general-purpose NLP library, Spacy\footnote{https://spacy.io}. Table~\ref{tab:features} shows the set of features we use for the CRF. 

The feature functions also include transition features that estimate the probability of a label, given the estimated probability of the preceding label. The transition features help the prediction of an implicit dependency, as the parse trees corresponding to the subsequent labels are disjoint, giving no evidence of the grammatical relation.
\begin{table}[]
	\centering
	\caption{Observation features used for predicting task dependency. The \textit{task\_type*} feature is optional and we show its efficacy in the evaluation section.}
	\begin{tabular}{|p{1.8cm}|p{5.4cm}|}
		\hline
		\textbf{Feature} & \textbf{Description} \\ \hline
		pos & Parts of speech tag of the token \\ \hline
		dep & Dependency relation from its parent in the tree \\ \hline
		has\_mark & True if the token has a child marking a subordinate clause \\ \hline
		advmod\_child & adverbial modifier of the token \\ \hline
		has\_advcl\_child & True if it has a adverbial clause modifier as a child \\ \hline
		length\_conj & No. of other tokens that are conjunctions of the token \\ \hline 
		task\_type* & Task type of the token \\ \hline
	\end{tabular}
	\label{tab:features}
\end{table}
\subsection{Task planning}
To execute a task, a robot needs to perform a sequence of low-level actions. A task plan is a sequence of such actions that satisfies the intended goal of the task. We consider a task specified in an instruction to change a hypothetical state of the world (initial state) to an expected state (goal state). We encode the initial and goal conditions of a task as a conjunction of fluents expressed in first-order logic. The templates are grounded using the predictions of the task interpreter to generate a planning problem in the PDDL formal language~\cite{mcdermott1998pddl}. During this grounding, if some necessary argument for the template is missing from the instruction, or it could not be predicted by the argument prediction model, our system asks the human for the same. For this dialogue, we resort to the dialogue strategy described in~\cite{pramanick2019strategy}. 

During the grounding of the templates, the assumed initial conditions for a task are updated by the post-conditions of the actions of the previous sequential task. In the case of conditionals, we generate a plan for each conditional-dependent pair, and in run-time, the correct action sequence is chosen from the actual observed outcome of the conditional task. Therefore, we reduce the problem of generating a task plan for the complex instruction, to the generation of a correct ordering of the tasks catering to the execution dependencies, followed by planning individually for the goals of the tasks in order while updating the assumed initial states using the post-conditions of the PDDL operators.

We deterministically order the predicted tasks to a control flow graph, organized as a tree, where each node in the tree denotes a task. An example of such a graph is illustrated in Fig.~\ref{fig:pipeline}. In this process, we make sure a conditional task is planned before any of its dependent tasks to resolve ordering dependency. To resolve execution dependency, we make a new branch when adding a new node, if it is dependent on the parent node. We add a dependent node to the left sub-tree if the dependency label is positive, and if the dependency label is negative, we add it to the right sub-tree. For tasks having a \textit{sequential} dependency label, we order them as per their corresponding appearances in the instruction.

In the case of multiple conditional tasks in the same instruction, we assume two such conditionals are indicating the same condition if the two tasks have the same type and have no dissimilar argument values. If so, we merge the two nodes and add the subsequent dependent tasks in the appropriate branches of the original conditional node. Otherwise, we consider the subsequent task to be a new conditional task and therefore make a new branch.

\section{EVALUATION}
\label{sec:eval}
We evaluate \system from multiple aspects -- (i)~compare our proposed approach for argument extraction against existing approaches, (ii)~report the performance of the CRF model for task dependency prediction, and (iii)~compare the end-to-end system performance of finding the required control flow graph of the given task against a baseline.

\subsection{Dataset}
As we do not assume a complex instruction all the time, we have considered a mixture of both simple and complex instructions. We have taken instructions containing a single task from the HuRIc dataset~\cite{bastianelli2014huric}. We have also created 182 additional samples of complex instruction with varying number of tasks per instruction (on average 3.73 tasks per instruction with s.d=1.93). This results in a dataset of 537 instructions in total. We trained the task and argument prediction models with this dataset using 80\% as training data and tested on the remaining 20\%. As we invoke the task dependency model only if there are multiple tasks, we trained the model for dependency prediction using 80\% of the complex instructions and tested its accuracy and end-to-end performance on the remaining 20\% complex instructions.

\subsection{Performance of task and argument prediction}
The CRF models for sequence labeling of task and argument types in \system, are trained with 8 different task type labels and 21 argument type labels. The task type prediction model has an F1-score of 0.94 on the test data. For the prediction of the argument type, we have compared our method with two other baseline methods. In the first baseline method, arguments are predicted solely with the linguistic feature without using the task type information. In the second baseline method, the task type information is naively associated with the argument predictor. In this case, it is assumed that the tokens followed by the task are part of its argument set. On the other hand, \system does have the strong assumption that tokens that follow the task are the only target for argument prediction. We find that the first baseline argument prediction model generates the same argument labeling only 67\% of the time, and generates F1 score of 0.82 (weighted average score of all argument labels) and the second baseline method achieves an exact match accuracy of 73\% with an F1 score of 0.91 (Table~\ref{tab:arg-performance}). This reveals the importance of associating the task type information for argument prediction, even if done naively. Our proposed model generates an exact match of 78\% and an F1 score of 0.93 on the test data. This improvement is due to correctly predicting arguments for statements in the instructions. Please note the model is not limited to the number of task and argument types (classes). The CRF can be trained and used with an additional or completely new set of task and argument types.
\begin{table}[]
	\centering
	\caption{Comparison of accuracy in predicting arguments.}
	\begin{tabular}{|l|l|l|}
		\hline
		\textbf{Method} & \textbf{Exact match} & \textbf{F1 score} \\ \hline
		Without task type association & 67\% & 0.82 \\ \hline
		Naive task type association & 72.7\% & 0.91 \\ \hline
		\system & 78.4\% & 0.93 \\ \hline 
	\end{tabular}
	\label{tab:arg-performance}
\end{table}

\subsection{Performance of task dependency resolver}
To evaluate the CRF model for task dependency labeling, we have compared our model with the ``lexicon induction'' rule based method (lex\_induct) as described in~\cite{misra2015environment} to predict the conditional and the dependent tasks. If none of the rules apply, the lex\_induct model predicts the label as \textit{sequential}. We consider two of our CRF models for the sequence labeling, one that uses all the features shown in Table~\ref{tab:features} and another that uses all the features except the optional \textit{task\_type} feature. 

We trained the models using the training data and report the accuracy metrics of our models and the lex\_induct on test data in Table~\ref{tab:dependecy-results}. The lex\_induct model achieves a F1-score of 0.76. The poor performance is attributed to inability to predict the dependency labels for out-of-order tasks and implicit dependencies. 
In comparison, our CRF model that doesn't use the \textit{task\_type} feature (\system\textsuperscript{-TF}), outperforms the lex\_induct by a large margin, $\delta F1$~=~+0.17. This is because our model uses both the syntactical relations and the transition features to predict the unmarked, implicit, and out-of order dependencies. For example, the transition features estimate higher likelihoods of the subsequent task of a conditional to have a positive or negative dependent label, while estimating low likelihoods for a subsequent task to have the \textit{sequential} label. This is revealed by the learnt weights of the transition features that are shown in Table~\ref{tab:transition}.


\begin{table}
	\centering
	\caption{Comparison of two of our proposed CRF models for task dependency prediction with a state-of-the-art (lex\_induct) method~\cite{misra2015environment}.}
	\begin{tabular}{|l|l|l|l|l|}
		\hline
		\textbf{Label} & \textbf{Model} & \textbf{Precision} & \textbf{Recall} & \textbf{F1} \\ \hline
		conditional & lex\_induct &0.74      &0.81      &0.77 \\ 
		& \system\textsuperscript{-TF} & 0.95      &0.86      &0.90   \\
		& \system &  0.91      &0.95      &0.93 \\ \hline
		dependent\_positive & lex\_induct &0.10      &0.07      &0.08 \\
		& \system\textsuperscript{-TF} &0.85      &0.79      &0.81 \\
		& \system & 0.87      &0.93      &0.90  \\ \hline
		dependent\_negative & lex\_induct &1.00      &0.45      &0.62 \\
		& \system\textsuperscript{-TF} &0.91      &0.91      &0.91 \\
		& \system &0.91      &0.91      &0.91 \\ \hline
		sequential & lex\_induct   &0.83      &0.89      &0.86 \\
		& \system\textsuperscript{-TF} &0.94      &0.97      &0.96 \\ 
		& \system &  0.98      &0.96      &0.97\\ \hline
		\color{blue}\textbf{Weighted average} & \color{blue}\textbf{lex\_induct} & \color{blue}\textbf{0.76} &\color{blue}\textbf{0.77} &\color{blue}\textbf{0.76}  \\
		& \color{blue}\textbf{\system\textsuperscript{-TF}} &\color{blue}\textbf{0.93}\ &\color{blue}\textbf{0.93} &\color{blue}\textbf{0.93} \\
		& \color{blue}\textbf{\system} &  \color{blue}\textbf{0.95}  & \color{blue}\textbf{0.95} & \color{blue}\textbf{0.95}  \\ \hline
	\end{tabular}
	\label{tab:dependecy-results}
\end{table}

\definecolor{aqua}{rgb}{0.0, 1.0, 1.0}
\definecolor{aquamarine}{rgb}{0.5, 1.0, 0.83}
\definecolor{ao}{rgb}{0.0, 0.0, 1.0}
\definecolor{blue}{rgb}{0.0, 0.0, 1.0}
\definecolor{blue(ryb)}{rgb}{0.01, 0.28, 1.0}
\definecolor{bisque}{rgb}{1.0, 0.89, 0.77} 	
\definecolor{bananamania}{rgb}{0.98, 0.91, 0.71}
\definecolor{bananayellow}{rgb}{1.0, 0.88, 0.21}
\definecolor{arylideyellow}{rgb}{0.91, 0.84, 0.42}
\definecolor{apricot}{rgb}{0.98, 0.81, 0.69}
\definecolor{aureolin}{rgb}{0.99, 0.93, 0.0}
\definecolor{awesome}{rgb}{1.0, 0.13, 0.32}
\definecolor{bittersweet}{rgb}{1.0, 0.44, 0.37}
\definecolor{dandelion}{rgb}{0.94, 0.88, 0.19}
\definecolor{buff}{rgb}{0.94, 0.86, 0.51}
\definecolor{cyan(process)}{rgb}{0.0, 0.72, 0.92}
\definecolor{cottoncandy}{rgb}{1.0, 0.74, 0.85}
\definecolor{corn}{rgb}{0.98, 0.93, 0.36}
\definecolor{flavescent}{rgb}{0.97, 0.91, 0.56}
\definecolor{babyblue}{rgb}{0.54, 0.81, 0.94}
\definecolor{beaublue}{rgb}{0.74, 0.83, 0.9}
\definecolor{blizzardblue}{rgb}{0.67, 0.9, 0.93}
\definecolor{columbiablue}{rgb}{0.61, 0.87, 1.0}


\begin{table*}[!t]
	\centering
	\caption{Weights of the transition features learnt by CRF. The weight values are used to calculate transition probabilities. The blue, yellow and pink cells denote a high, medium and low transition likelihood, respectively. }
	\begin{tabular}{|>{\centering\arraybackslash}p{3cm}|>{\centering\arraybackslash}p{3cm}|>{\centering\arraybackslash}p{3cm}|>{\centering\arraybackslash}p{3cm}|>{\centering\arraybackslash}p{3cm}|}
		\hline
		\textbf{Previous prediction} &\textbf{condition}    &\textbf{dependent\_negative}    &\textbf{dependent\_positive}    &\textbf{sequential} \\ \hline
		condition &\cellcolor{flavescent}0.001 &\cellcolor{columbiablue}6.464   &\cellcolor{columbiablue}5.346  &\cellcolor{cottoncandy}-1.339 \\ \hline
		dependent\_positive  &\cellcolor{flavescent}0.196   &\cellcolor{columbiablue}4.280   &\cellcolor{columbiablue}1.500  &\cellcolor{cottoncandy}-2.593 \\ \hline
		dependent\_negative  &\cellcolor{cottoncandy}-3.105 &\cellcolor{flavescent}0.001   &\cellcolor{flavescent}0.001   &\cellcolor{flavescent}-0.692 \\ \hline
		sequential &\cellcolor{columbiablue}3.018 &\cellcolor{columbiablue}4.020  &\cellcolor{cottoncandy}-3.801 &\cellcolor{columbiablue}2.535 \\ \hline
	\end{tabular}
	\label{tab:transition}
\end{table*}

Using the \textit{task\_type} feature further improves the performance of the CRF, showing an  overall improvement of $\delta F1$~=~+0.02 over the CRF that doesn't use the feature. This is because this model associates high probabilities of the \textit{conditional} label with certain types of tasks that are often used to express a conditional task, such as searching for an object or checking the state of an object. Although, if this co-relation of a task and its dependency type is not present in an application domain, still our model that does not use the \textit{task\_type} feature can be used that has an acceptable accuracy.

\subsection{End-to-end performance}
\begin{figure*}
	\centering
	\subfloat[exact match accuracy]{\includegraphics[width=0.45\textwidth]{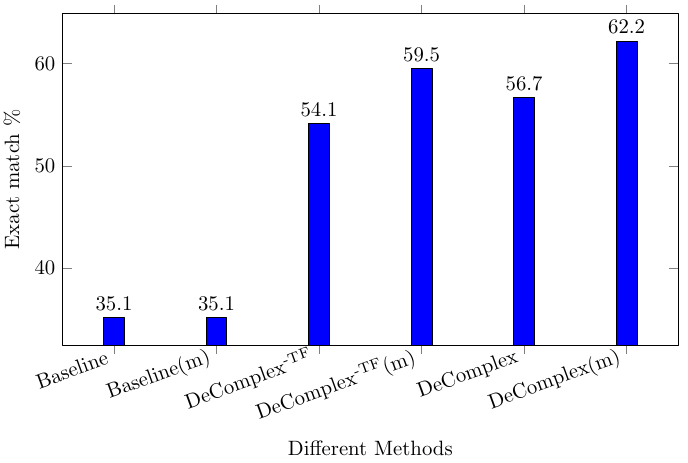}\label{fig:exact-match-end}}
	\subfloat[ordering error rate]{\includegraphics[width=0.45\textwidth]{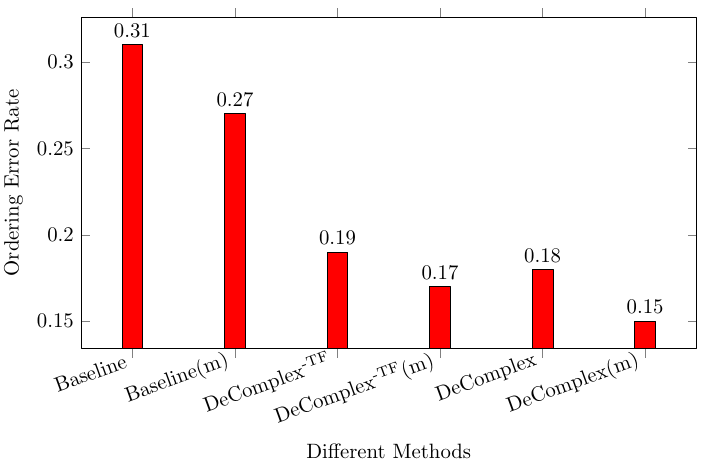}\label{fig:oer}}
	\caption{Comparison of our approaches against a baseline in terms of exact match and ordering error rate. A method marked with a `\textit{(m)}' denotes that it uses our strategy to merge redundant tasks.}
	\label{fig:end-to-end}
\end{figure*}

We have evaluated the end-to-end performance of \system to find the intended control flow graph of tasks, from a natural language instruction, using two metrics. Firstly, we have calculated the number of exact matches between a graph generated by our system and the corresponding annotated graph. We annotated the correct control flow graphs for all the complex instructions in test data considering their task and dependency types and merged redundant nodes (tasks) if any. For task identification part, we used our own CRF model in all the three system variants, as the lex\_induct model doesn't predict task types. We show the result of the comparisons in Fig.~\ref{fig:exact-match-end}.  

We find that lex\_induct system performs poorly, as it can find the exactly same graph only 35\% of the time. Even after using our technique of merging it gives the same percentage of exact match. This is because can't predict the execution and ordering dependency for many examples, therefore does not benefit from the merging strategy. We see a 19\% improvement in performance by using our CRF model for dependency resolution, even without using the task type information and merging strategy. When we merge redundant tasks along with this model, we generate the correct graph 60\% of the time. Subsequently, when we use our full CRF model for dependency resolving, but do not use merging, we get an exact match accuracy of 57\%. Interestingly, even though the dependency resolving CRF is better trained, its accuracy is lesser than the model that that uses a relatively weaker dependency resolver, but merges the redundant tasks. This shows that the technique of merging redundant tasks is certainly useful when handling long sequences of tasks in a complex instruction. Our full model finds an exact match 62\% of the time, outperforming the baseline by a large margin. 

As the task identifier is probabilistic, its error propagates to the dependency resolver, i.e the predicted graph of tasks can deviate from the ground truth even if one single task is mis-predicted and even when the mis-predicted task is sequential. For this, we also use a less pessimistic metric, named as the Ordering Error Rate (OER). We use this metric by following by a similar metric used in~\cite{misra2015environment,misra2016tell}. We define OER as the number of Substitutions (S), Deletion (D) and Insertion (I) of nodes performed on the predicted control flow graph to produce the ground truth graph, divided by the number of nodes in ground truth, i.e., for a ground truth graph of $N$ tasks,
\[ \small OER = \frac{S+D+I}{N} .\]

We show the results of the OER metric for the baseline system and \system along with their corresponding redundant task merging variants, in Fig.~\ref{fig:oer}. By analyzing the results, we find that the control flow graphs generated by our full model is very similar to the ground truth (OER=0.15) and the error slightly increases (OER=0.18) when redundant tasks are not merged. These error rates are closely followed by \system\textsuperscript{-TF}, which receives an error rate of 0.19 before and 0.17 after merging. Whereas for the lex\_induct system, the predicted graphs largely differ from ground truth (OER=0.31), even though it uses the same task identification model. The error is slightly reduced (OER=0.27) by using our merging strategy.
\subsection{Discussion}
By analyzing the failure cases, we see that the decline in end-to-end performance from the individual accuracy of the task dependency resolver is mainly attributed to the errors made by the task identification model, whose mis-prediction of a single task leads to an in-exact match. This particularly the case when the instructions are very long, containing more than 5 tasks. There are also a few cases where the task is predicted correctly, but the dependency label is incorrect, leading to a node insertion in an incorrect branch. However, the limitations of \system can be overcome by a suitable dialogue agent that can ask appropriate questions to correct the individual mis-predictions. In future, we plan to extend \system by integrating a dialogue engine with it. In this work we take textual input instead of verbal instruction, assuming an accurate speech to text conversion system. Mitigating the propagation of error by a noisy speech transcription, specially in the case of a mobile robot and far-field speech, can be a valuable addition to the system.

\section{CONCLUSIONS}
\label{sec:con}
Providing instructions to a robot through natural language conversation adds to the usability of the robot and convenience for the user. The instructions are often provided as a complex phrase, especially when neither the user nor the robot has a full view of the environment. Existing work often assumes simple task instructions with a single task or multiple independent tasks. However, when multiple tasks are present in such a complex instruction, it includes situations where the execution of certain tasks are dependent on the outcome of another. Most of the time, such an inter-dependency between tasks is not stated explicitly, which makes its prediction a challenging task. In this work, we have presented a method to understand such dependencies between tasks and re-order the tasks catering to their dependency types. We have presented a probabilistic model and pointed out the useful features to predict the dependencies with high accuracy. After finding the required order of task execution, we plan for each task in the order after merging redundant tasks (if any) and generate conditional plans for the dependent tasks. We have compared our system by designing a baseline based on existing work and found that our system significantly outperforms the existing system.

\balance
\bibliographystyle{IEEEtran}
\bibliography{main}

\begin{thebibliography}{10}
\providecommand{\url}[1]{#1}
\csname url@samestyle\endcsname
\providecommand{\newblock}{\relax}
\providecommand{\bibinfo}[2]{#2}
\providecommand{\BIBentrySTDinterwordspacing}{\spaceskip=0pt\relax}
\providecommand{\BIBentryALTinterwordstretchfactor}{4}
\providecommand{\BIBentryALTinterwordspacing}{\spaceskip=\fontdimen2\font plus
\BIBentryALTinterwordstretchfactor\fontdimen3\font minus
  \fontdimen4\font\relax}
\providecommand{\BIBforeignlanguage}[2]{{%
\expandafter\ifx\csname l@#1\endcsname\relax
\typeout{** WARNING: IEEEtran.bst: No hyphenation pattern has been}%
\typeout{** loaded for the language `#1'. Using the pattern for}%
\typeout{** the default language instead.}%
\else
\language=\csname l@#1\endcsname
\fi
#2}}
\providecommand{\BIBdecl}{\relax}
\BIBdecl

\bibitem{pramanick2018defatigue}
P.~Pramanick and C.~Sarkar, ``Defatigue: Online non-intrusive fatigue detection
  by a robot co-worker,'' in \emph{2018 27th IEEE International Symposium on
  Robot and Human Interactive Communication (RO-MAN)}.\hskip 1em plus 0.5em
  minus 0.4em\relax IEEE, 2018, pp. 1129--1136.

\bibitem{martinez2019pharos}
E.~Martinez-Martin, A.~Costa, and M.~Cazorla, ``Pharos 2.0—a physical
  assistant robot system improved,'' \emph{Sensors}, vol.~19, no.~20, p. 4531,
  2019.

\bibitem{shah2018follownet}
P.~Shah, M.~Fiser, A.~Faust, C.~Kew, and D.~Hakkani-Tur, ``Follownet: Robot
  navigation by following natural language directions with deep reinforcement
  learning,'' in \emph{Third Machine Learning in Planning and Control of Robot
  Motion Workshop at ICRA}, 2018.

\bibitem{nicolescu2019learning}
M.~Nicolescu, N.~Arnold, J.~Blankenburg, D.~Feil-Seifer, S.~B. Banisetty,
  M.~Nicolescu, A.~Palmer, and T.~Monteverde, ``Learning of complex-structured
  tasks from verbal instruction,'' in \emph{International Conference on
  Humanoid Robots}, Toronto, Canada, October 2019.

\bibitem{lu2017integrating}
D.~Lu, Y.~Zhou, F.~Wu, Z.~Zhang, and X.~Chen, ``Integrating answer set
  programming with semantic dictionaries for robot task planning,'' in
  \emph{Proceedings of the 26th International Joint Conference on Artificial
  Intelligence}.\hskip 1em plus 0.5em minus 0.4em\relax AAAI Press, 2017, pp.
  4361--4367.

\bibitem{paul2018temporal}
R.~Paul, A.~Barbu, S.~Felshin, B.~Katz, and N.~Roy, ``Temporal grounding graphs
  for language understanding with accrued visual-linguistic context,'' in
  \emph{Proceedings of the 26th International Joint Conference on Artificial
  Intelligence}.\hskip 1em plus 0.5em minus 0.4em\relax AAAI Press, 2017, pp.
  4506--4514.

\bibitem{matuszek2013learning}
C.~Matuszek, E.~Herbst, L.~Zettlemoyer, and D.~Fox, ``Learning to parse natural
  language commands to a robot control system,'' in \emph{Experimental
  Robotics}.\hskip 1em plus 0.5em minus 0.4em\relax Springer, 2013, pp.
  403--415.

\bibitem{tellex2011understanding}
S.~Tellex, T.~Kollar, S.~Dickerson, M.~R. Walter, A.~G. Banerjee, S.~Teller,
  and N.~Roy, ``Understanding natural language commands for robotic navigation
  and mobile manipulation,'' in \emph{Twenty-Fifth AAAI Conference on
  Artificial Intelligence}, 2011.

\bibitem{pramanick2019enabling}
P.~Pramanick, C.~Sarkar, P.~Balamuralidhar, A.~Kattepur, I.~Bhattacharya, and
  A.~Pal, ``Enabling human-like task identification from natural
  conversation,'' in \emph{2019 IEEE/RSJ International Conference on
  Intelligent Robots and Systems (IROS)}.\hskip 1em plus 0.5em minus
  0.4em\relax IEEE, 2019.

\bibitem{thomas2012roboframenet}
B.~J. Thomas and O.~C. Jenkins, ``Roboframenet: Verb-centric semantics for
  actions in robot middleware,'' in \emph{2012 IEEE International Conference on
  Robotics and Automation}.\hskip 1em plus 0.5em minus 0.4em\relax IEEE, 2012,
  pp. 4750--4755.

\bibitem{misra2015environment}
D.~K. Misra, K.~Tao, P.~Liang, and A.~Saxena, ``Environment-driven lexicon
  induction for high-level instructions,'' in \emph{Proceedings of the 53rd
  Annual Meeting of the Association for Computational Linguistics and the 7th
  International Joint Conference on Natural Language Processing (Volume 1: Long
  Papers)}, 2015, pp. 992--1002.

\bibitem{antunes2016human}
A.~Antunes, L.~Jamone, G.~Saponaro, A.~Bernardino, and R.~Ventura, ``From human
  instructions to robot actions: Formulation of goals, affordances and
  probabilistic planning,'' in \emph{Robotics and Automation (ICRA), 2016 IEEE
  International Conference on}.\hskip 1em plus 0.5em minus 0.4em\relax IEEE,
  2016, pp. 5449--5454.

\bibitem{chen2013handling}
X.-P. Chen, J.-M. Ji, Z.-Q. Sui, and J.-k. Xie, ``Handling open knowledge for
  service robots,'' in \emph{Twenty-Third International Joint Conference on
  Artificial Intelligence}, 2013.

\bibitem{nyga2018grounding}
D.~Nyga, S.~Roy, R.~Paul, D.~Park, M.~Pomarlan, M.~Beetz, and N.~Roy,
  ``Grounding robot plans from natural language instructions with incomplete
  world knowledge,'' in \emph{Proceedings of The 2nd Conference on Robot
  Learning}, ser. Proceedings of Machine Learning Research, vol.~87, 2018, pp.
  714--723.

\bibitem{thomason2015learning}
J.~Thomason, S.~Zhang, R.~J. Mooney, and P.~Stone, ``Learning to interpret
  natural language commands through human-robot dialog.'' in \emph{IJCAI},
  2015, pp. 1923--1929.

\bibitem{thomason2019improving}
J.~Thomason, A.~Padmakumar, J.~Sinapov, N.~Walker, Y.~Jiang, H.~Yedidsion,
  J.~Hart, P.~Stone, and R.~J. Mooney, ``Improving grounded natural language
  understanding through human-robot dialog,'' in \emph{International Conference
  on Robotics and Automation (ICRA)}.\hskip 1em plus 0.5em minus 0.4em\relax
  IEEE, 2019.

\bibitem{pramanick2019strategy}
P.~Pramanick, C.~Sarkar, and I.~Bhattacharya, ``Your instruction may be crisp,
  but not clear to me!'' in \emph{2019 IEEE International Conference on Robot
  and Human Interactive Communication (RO-MAN)}.\hskip 1em plus 0.5em minus
  0.4em\relax IEEE, 2019.

\bibitem{kollar2014grounding}
T.~Kollar, S.~Tellex, D.~Roy, and N.~Roy, ``Grounding verbs of motion in
  natural language commands to robots,'' in \emph{Experimental robotics}.\hskip
  1em plus 0.5em minus 0.4em\relax Springer, 2014, pp. 31--47.

\bibitem{kollar2010toward}
------, ``Toward understanding natural language directions,'' in \emph{2010 5th
  ACM/IEEE International Conference on Human-Robot Interaction (HRI)}.\hskip
  1em plus 0.5em minus 0.4em\relax IEEE, 2010, pp. 259--266.

\bibitem{kress2008translating}
H.~Kress-Gazit, G.~E. Fainekos, and G.~J. Pappas, ``Translating structured
  english to robot controllers,'' \emph{Advanced Robotics}, vol.~22, no.~12,
  pp. 1343--1359, 2008.

\bibitem{mericcli2014interactive}
C.~Meri{\c{c}}li, S.~D. Klee, J.~Paparian, and M.~Veloso, ``An interactive
  approach for situated task specification through verbal instructions,'' in
  \emph{Proceedings of the 2014 international conference on Autonomous agents
  and multi-agent systems}.\hskip 1em plus 0.5em minus 0.4em\relax
  International Foundation for Autonomous Agents and Multiagent Systems, 2014,
  pp. 1069--1076.

\bibitem{dzifcak2009and}
J.~Dzifcak, M.~Scheutz, C.~Baral, and P.~Schermerhorn, ``What to do and how to
  do it: Translating natural language directives into temporal and dynamic
  logic representation for goal management and action execution,'' in
  \emph{2009 IEEE International Conference on Robotics and Automation}.\hskip
  1em plus 0.5em minus 0.4em\relax IEEE, 2009, pp. 4163--4168.

\bibitem{finucane2010ltlmop}
C.~Finucane, G.~Jing, and H.~Kress-Gazit, ``Ltlmop: Experimenting with
  language, temporal logic and robot control,'' in \emph{2010 IEEE/RSJ
  International Conference on Intelligent Robots and Systems}.\hskip 1em plus
  0.5em minus 0.4em\relax IEEE, 2010, pp. 1988--1993.

\bibitem{chai2017teaching}
J.~Y. Chai, M.~Cakmak, and C.~Sidner, ``Teaching robots new tasks through
  natural interaction,'' in \emph{Interactive Task Learning: Agents, Robots,
  and Humans Acquiring New Tasks through Natural Interactions, Str{\"u}ngmann
  Forum Reports, J. Lupp, series editor}, vol.~26, 2017.

\bibitem{cantrell2012tell}
R.~Cantrell, K.~Talamadupula, P.~Schermerhorn, J.~Benton, S.~Kambhampati, and
  M.~Scheutz, ``Tell me when and why to do it!: Run-time planner model updates
  via natural language instruction,'' in \emph{Proceedings of the seventh
  annual ACM/IEEE international conference on Human-Robot Interaction}.\hskip
  1em plus 0.5em minus 0.4em\relax ACM, 2012, pp. 471--478.

\bibitem{boteanu2016model}
A.~Boteanu, T.~Howard, J.~Arkin, and H.~Kress-Gazit, ``A model for verifiable
  grounding and execution of complex natural language instructions,'' in
  \emph{2016 IEEE/RSJ International Conference on Intelligent Robots and
  Systems (IROS)}.\hskip 1em plus 0.5em minus 0.4em\relax IEEE, 2016, pp.
  2649--2654.

\bibitem{pomarlan2017natural}
M.~Pomarlan, S.~Koralewski, and M.~Beetz, ``From natural language instructions
  to structured robot plans,'' in \emph{Joint German/Austrian Conference on
  Artificial Intelligence (K{\"u}nstliche Intelligenz)}.\hskip 1em plus 0.5em
  minus 0.4em\relax Springer, 2017, pp. 344--351.

\bibitem{artzi2013weakly}
Y.~Artzi and L.~Zettlemoyer, ``Weakly supervised learning of semantic parsers
  for mapping instructions to actions,'' \emph{Transactions of the Association
  for Computational Linguistics}, vol.~1, pp. 49--62, 2013.

\bibitem{misra2016tell}
D.~K. Misra, J.~Sung, K.~Lee, and A.~Saxena, ``Tell me dave: Context-sensitive
  grounding of natural language to manipulation instructions,'' \emph{The
  International Journal of Robotics Research}, vol.~35, no. 1-3, pp. 281--300,
  2016.

\bibitem{mcdermott1998pddl}
M.~Ghallab, A.~Howe, C.~Knoblock, D.~McDermott, A.~Ram, M.~Veloso, D.~Weld, and
  D.~Wilkins, ``Pddl-the planning domain definition language,'' \emph{AIPS-98
  planning committee}, vol.~3, p.~14, 1998.

\bibitem{bastianelli2014huric}
E.~Bastianelli, G.~Castellucci, D.~Croce, L.~Iocchi, R.~Basili, and D.~Nardi,
  ``Huric: a human robot interaction corpus,'' in \emph{Proceedings of the
  Ninth International Conference on Language Resources and Evaluation
  (LREC-2014)}, 2014.

\end{thebibliography}

\end{document}